\documentclass[runningheads]{llncs}
\usepackage{graphicx}
\usepackage{comment}
\usepackage{amsmath,amssymb} 
\usepackage{color}
\usepackage{multirow}
\usepackage{csquotes}
\usepackage{rotating}
\usepackage{algorithm}
\usepackage{algorithmicx}
\usepackage{algpseudocode}
\usepackage{arydshln}
\usepackage{booktabs}

\newcommand\eg{\emph{e.g.}} 
\newcommand\ie{\emph{i.e.}} 
 
\newcommand\etc{\emph{etc.}}

\begin{document}
\pagestyle{headings}
\mainmatter
\def\ECCVSubNumber{2760}  

\title{Learning Noise-Aware Encoder-Decoder from Noisy Labels by Alternating Back-Propagation for Saliency Detection} 

\titlerunning{Noise-Aware Encoder-Decoder for Saliency Detection}

%
\author{Jing Zhang\inst{1,3,4}\thanks{Work was done while Jing Zhang was an intern mentored by Jianwen Xie.} \and
Jianwen Xie\inst{2} \and
Nick Barnes\inst{1}}
%
%
\institute{Australian National University, Australia \and
Cognitive Computing Lab, Baidu Research, USA \and
Australian Centre for Robotic Vision, Australia \and
Data61, Australia}
\maketitle

\begin{abstract}
In this paper, we propose a noise-aware encoder-decoder framework to disentangle a clean saliency predictor from noisy training examples, where the noisy labels are generated by
   unsupervised handcrafted feature-based methods.
   The proposed model consists of two sub-models parameterized by neural networks: (1) a saliency predictor that maps input images to clean saliency maps, and (2) a noise generator, which is a latent variable model that produces noises from  Gaussian latent vectors. The whole model that represents noisy labels is a sum of the two sub-models. The goal of training the model is to estimate the parameters of both sub-models, and simultaneously infer the corresponding latent vector of each noisy label. We propose to train the model by using an alternating back-propagation (ABP) algorithm, which alternates the following two steps: (1) learning back-propagation for estimating the parameters of two sub-models by gradient ascent, and (2) inferential back-propagation for inferring the latent vectors of training noisy examples by Langevin Dynamics. To prevent the network from converging to trivial solutions, we utilize an edge-aware smoothness loss to regularize hidden saliency maps to have similar structures as their corresponding images. Experimental results on several benchmark datasets indicate the effectiveness of the proposed model. 
\end{abstract}

\section{Introduction}

\begin{figure}[!htp]
   \begin{center}
   {\includegraphics[width=0.88\linewidth]{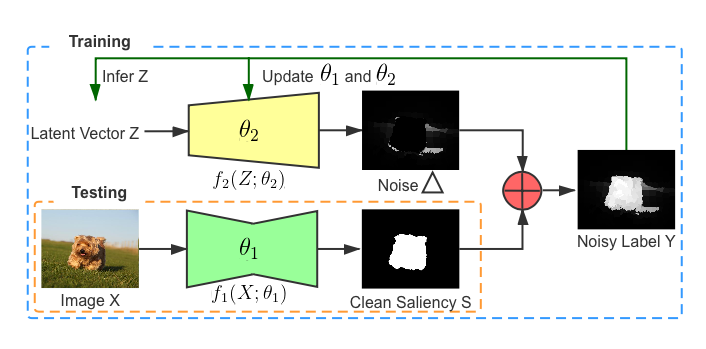}}
   \end{center}
   \caption{An
   illustration of our
   framework. Representation: Each noisy label $Y$ is represented as a sum of a clean saliency $S$ and a noise map $\Delta$. The clean saliency $S$ is predicted from an image $X$ by an encoder-decoder network $f_1$, and the noise is produced from a Gaussian noise vector $Z$ by a generator network $f_2$. Training: given the observed image $X$ and the corresponding noisy label $Y$, (i) the latent vector $Z$ is inferred by MCMC and (ii) the parameters $\{\theta_1, \theta_2\}$ of the encoder-decoder $f_1$ and the generator $f_2$ are updated by the gradient ascent for maximum likelihood. Testing: once the model is learned, the disentangled salicey predictor $f_1$ is the desired model for salicey prediction.
   }
   
   \label{fig:illustration}
\end{figure}
Visual saliency detection aims to locate
salient regions
that attract human attention.
Conventional saliency detection methods \cite{Background-Detection:CVPR-2014,Manifold-Ranking:CVPR-2013} rely on human designed features to compute saliency
for
each pixel or superpixel.
The deep learning revolution makes it possible to train end-to-end deep saliency detection models in a data-driven manner \cite{picanet,Amulet,CPD_Sal,prpgressive_attention,MSNet_Sal,AFNet_Sal,BASNet_Sal,PFPN_aaai2020,F3Net_aaai2020,Liu_2019_ICCV,Iter_Coop_CVPR,PAGE_cvpr19,fix_driven_sal,Zhang_2020_ucnet},
outperforming handcrafted feature-based solutions by a wide margin.
However, the success of deep models mainly depends on a large amount of accurate human labeling \cite{imagesaliency,MSRA10K,Dongxu_2020_WACV}, which is typically
expensive and time-consuming. 

To relieve the burden of pixel-wise labeling, weakly supervised \cite{Guanbin_weaksalAAAI,imagesaliency,scribble_jing} and unsupervised saliency detection models \cite{Zhang_2018_CVPR,Zhang_2017_ICCV,DeepUSPSDR} have been proposed. The former direction focuses on learning saliency from cheap but clean annotations,
while the latter one studies
learning saliency from noisy
labels, which are typically obtained by conventional handcrafted feature-based methods.
In this paper, we follow the second direction and propose a deep latent variable model that we call the noise-aware encoder-decoder to disentangle a clean saliency predictor from noisy labels. In general, a noisy label can be (1) a coarse saliency label generated by algorithmic pipelines using handcrafted features, (2) an imperfect human-annotated saliency label, or even (3) a clean label, which actually is a special case of noisy label, in which noise is none. 
Aiming at unsupervised saliency prediction, our paper assumes noisy labels to be produced by unsupervised handcrafted feature-based saliency  methods, and places emphasis on disentangled representation of noisy labels by the noise-aware encoder-decoder.  

Given a noisy dataset $D=\{(X_i,Y_i)\}_{i=1}^n$
of $n$ examples, where $X$
and $Y$ are image and its
corresponding noisy saliency label,
we intend to disentangle noise $\Delta$ and clean saliency $S$ from each noisy label $Y$, and learn a clean saliency predictor $f_1: X \rightarrow S$. To achieve this, we propose a conditional latent variable model, which is a disentangled representation of noisy saliency $Y$.
See Figure \ref{fig:illustration} for an illustration of the proposed model.
In the context of the model, each noisy label is assumed to be generated by adding a specific noise or perturbation $\Delta$ to its clean saliency map $S$ that is dependent on its image $X$.
Specifically, the model consists of two sub-models: (1) saliency predictor $f_1$: an encoder-decoder network that maps an input image $X$ to a latent clean saliency map $S$, and (2)  noise generator $f_2$: a top-down neural network that produces a noise or error $\Delta$ from a low-dimensional Gaussian latent vector $Z$. 

As a latent variable model, the rigorous maximum likelihood learning (MLE) typically requires to compute an intractable posterior distribution, which is an inference step.
To learn the latent variable model, two algorithms can be adopted: variational auto-encoder (VAE) \cite{VAE_2014} or alternating back-propagation (ABP) \cite{ABP_AAAI,xie2019learning,zhu2019learning}. VAE approximates MLE by minimizing the evidence lower bound with a separate inference model to approximate the true posterior, while ABP directly targets MLE and computes the posterior via Markov chain Monte Carlo (MCMC).
In this paper, we generalize the ABP algorithm
to learn the proposed model, which alternates the following two steps: (1) learning back-propagation for estimating the parameters of two sub-models, and (2) inferential back-propagation for inferring the latent vectors of training examples. As there may exist infinite combinations of $S$ and $\Delta$ such that $S+\Delta$ perfectly matches the provided noisy label $Y$, we further adopt the edge-aware smoothness loss \cite{occlusion_aware} to serve as a regularization to force each latent saliency map $S$ to have a similar structure as its input image $X$. The learned disentangled saliency predictor $f_1$ is the desired model for testing.

Our solution is different from existing weak or noisy label-based saliency approaches \cite{Zhang_2018_CVPR,Zhang_2017_ICCV,DeepUSPSDR,xin2018c2s} in the following aspects:
Firstly, unlike \cite{Zhang_2018_CVPR}, we don't assume the saliency noise distribution is a Gaussian distribution. Our noise generator parameterized by a neural network is flexible enough to approximate any forms of structural noises.
Secondly, we design a trainable noise generator to  explicitly represent each noise $\Delta$ as a non-linear transformation of low-dimensional Gaussian noise $Z$, which is a latent variable that need to be inferred during training, while \cite{Zhang_2018_CVPR,Zhang_2017_ICCV,DeepUSPSDR,xin2018c2s} have no noise inference process. 
Thirdly, we have no constraints on the number of noisy labels generated from each image, while \cite{Zhang_2018_CVPR,Zhang_2017_ICCV,DeepUSPSDR}
require multiple noisy labels per image for noise modeling or pseudo label generation.
Lastly, our edge-aware smoothness loss serves as a regularization to force the produced latent saliency maps to be well aligned with their input images, which is different from \cite{xin2018c2s}, where object edges are used to produce pseudo saliency labels
via multi-scale combinatorial grouping (MCG) \cite{MCG}.

Our main contributions can be summarized as follows:
\begin{itemize}
    \item We propose to learn a clean saliency predictor from noisy labels by a novel latent variable model that we call noise-aware encoder-decoder, in which each noisy label is represented as a sum of the clean saliency generated from the input image and a noise map generated from a latent vector. 
    \item We propose to train the proposed model by an alternating back-propagation (ABP) algorithm, which rigorously and efficiently maximizes the data likelihood without recruiting any other auxiliary model. 
    \item We propose to use an edge-aware smoothness loss as a regularization to prevent the model from converging to a trivial solution.
    \item Experimental results on various benchmark datasets show the state-of-the-art performances of our framework in the task of unsupervised saliency detection, and also comparable performances with the existing fully-supervised saliency detection methods. 
\end{itemize}

\section{Related Work}

\textbf{Fully supervised saliency detection models} \cite{PFPN_aaai2020,F3Net_aaai2020,Liu_2019_ICCV,Iter_Coop_CVPR,PAGE_cvpr19,BASNet_Sal,CPD_Sal,pyramid_attention_edge,zhao2019EGNet,zhao2019Contrast} mainly focus on designing networks that utilize image context information, multi-scale information, and image structure preservation. \cite{PFPN_aaai2020} introduces feature polishing modules to
update each level of features by incorporating all higher levels of context information.
\cite{F3Net_aaai2020} presents a cross feature module and a cascaded feedback decoder to effectively fuse different levels of features with a position-aware loss to penalize the boundary as well as pixel dissimilarity between saliency outputs and labels during training.
\cite{Iter_Coop_CVPR}
proposes a saliency detection model that integrates both top-down and bottom-up saliency inferences in an iterative and cooperative manner. \cite{PAGE_cvpr19} designs a pyramid attention structure with an edge detection module to perform edge-preserving salient object detection.
\cite{BASNet_Sal} uses a hybrid loss for boundary-aware saliency detection.
\cite{pyramid_attention_edge} proposes to use the stacked pyramid attention, which exploits multi-scale saliency information, along with an edge-related loss for saliency detection. 

\textbf{Learning saliency models without pixel-wise labeling } can relieve the burden of costly pixel-level labeling. Those methods  train saliency detection models with low-cost labels, such as
image-level labels \cite{imagesaliency,Guanbin_weaksalAAAI,MSW_Sal}, noisy labels \cite{Zhang_2018_CVPR,Zhang_2017_ICCV,DeepUSPSDR}, object contours \cite{xin2018c2s}, scribble annotations \cite{scribble_jing}, \etc 
\cite{imagesaliency} introduces a foreground inference network to produce initial saliency maps with image-level labels, which are further refined
and then treated as pseudo labels for iterative training.
\cite{Zhang_2017_ICCV} fuses saliency maps from unsupervised handcrafted feature-based methods with heuristics within a deep learning framework. 
\cite{Zhang_2018_CVPR}
collaboratively updates a saliency prediction module and a noise module to achieve learning saliency from multiple noisy labels. 
In \cite{DeepUSPSDR}, the initial noisy labels
are refined by
a self-supervised learning technique, and then
treated as pseudo labels.
\cite{xin2018c2s} creates a contour-to-saliency network, where
saliency masks are generated by its contour detection branch via MCG \cite{MCG} and then those generated saliency masks are further used to train its saliency detection branch.

\textbf{Learning from noisy labels} techniques mainly focus on
three main directions:
(1) developing regularization \cite{Reed2014TrainingDN,Yi_2019_CVPR}; 
(2) estimating the noise distribution by assuming that noisy labels are corrupted from clean labels by an unknown noise transition matrix~\cite{Noiseadapt,Tanno2019LearningFN} and 
(3) training on selected samples~\cite{mentornet,importance_reweighting}. \cite{Reed2014TrainingDN} deals with noisy labeling by augmenting the prediction objective with a notion of perceptual consistency.
\cite{Yi_2019_CVPR} proposes a framework to solve noisy label problem by updating both model parameters and labels.
\cite{Tanno2019LearningFN} proposes to
simultaneously learn the individual annotator model, which is represented by a confusion matrix, and
the underlying true label distribution (\ie, classifier) from noisy observations. 
\cite{mentornet} proposes to learn an extra network called MentorNet to generate a curriculum, which is a sample weighting scheme, for the base ConvNet called StudentNet. The generated curriculum helps the StudentNet to focus on those samples whose labels are likely to be correct.

\section{Proposed Framework}

The proposed model consists of two sub-models:
(1) a saliency predictor, which is parameterized by an encoder-decoder network that maps the input image $X$ to the clean saliency $S$;
(2) a noise generator, which is parameterized by a top-down generator network that produces
a noise or error $\Delta$ from a Gaussian latent vector $Z$. The resulting model is a sum of the two sub-models. Given training images with noisy labels, the MLE training of the model leads to an alternating back-propagation algorithm, which will be introduced in details in the following sections.
The learned encoder-decoder network, which takes as input an image $X$ and outputs its clean saliency $S$, is the disentangled model for saliency detection. 

\subsection{Noise-Aware Encoder-Decoder Network}
\label{problem_formulation}
Let $D = \{(X_i,Y_i)\}_{i=1}^n$ be the training dataset,
where $X$ is the
training image, $Y$ is the noisy label of $X$, $n$ is the size of the training dataset. Formally, the noise-aware encoder-decoder model can be formulated as follows:
\begin{eqnarray}
    && S= f_{1}(X; \theta_1), \label{eq:1}\\
    && \Delta = f_{2}(Z; \theta_2), Z \sim \mathcal{N}(0,I_d), \label{eq:2}  \\
    && Y= S+ \Delta + \epsilon, \epsilon \sim \mathcal{N}(0,\sigma^2 I_D),  \label{eq:3} 
\end{eqnarray}
\noindent
where $f_{1}$ in Eq.~(\ref{eq:1}) is an encoder-decoder structure parameterized by $\theta_1$ for saliency detection. It takes  as input an image $X$ and predicts its clean saliency map $S$.
Eq. (\ref{eq:2}) defines a noise generator, where $Z$ is a low-dimensional Gaussian noise vector following $\mathcal{N}(0,I_d)$ ($I_d$ is the $d$-dimensional identity matrix) and $f_{2}$ is a top-down deconvolutional neural network parametrized by $\theta_2$ that generates a saliency noise $\Delta$ from the noise vector $Z$. 
In Eq. (\ref{eq:3}), we assume that the observed noisy label $Y$ is a sum of the clean saliency map $S$ and the noise $\Delta$, plus a Gaussian residual $\epsilon \sim \mathcal{N}(0, \sigma^2 I_D)$, where we assume $\sigma$ is given and $I_D$ is the $D$-dimensional identity matrix.
Although $Z$ is a Gaussian noise, the generated noise $\Delta$ is not necessarily Gaussian due to the non-linear transformation $f_2$.

We call our network the noise-aware encoder-decoder network as it explicitly decomposes a noisy label $Y$ into a noise $\Delta$ and a clean label $S$, and simultaneously learns a mapping from the image $X$ to the clean saliency map $S$ via an encoder-decoder network as shown in Fig. \ref{fig:illustration}. Since the resulting model involves latent variables $Z$, training the model by maximum likelihood learning typically needs to learn the parameters $\theta_1$ and $\theta_2$, and also infer the noise latent variable $Z_i$ for each observed data pair $(X_i, Y_i)$. The noise and the saliency information are disentangled once the model is learned. The learned encoder-decoder sub-model $S=f_1(X; \theta_1)$ is the desired saliency detection network.

\subsection{Maximum Likelihood via Alternating Back-Propagation}
For notation simplicity, let $f=\{f_1, f_2\}$ and $\theta=\{\theta_1, \theta_2\}$. The proposed model is rewritten as a summarized form: $Y=f(X, Z; \theta)+\epsilon$, where $Z \sim \mathcal{N}(0,I_d)$ and $\epsilon$ is the observation error. Given a dataset $D=\{(X_i,Y_i)\}_{i=1}^n$, each training example $(X_i,Y_i)$ should have a corresponding $Z_i$, but all data shares the same model parameter $\theta$. Intuitively, we should infer $Z_i$ and learn $\theta$ to minimize the reconstruction error $\sum_{i=1}^n \|Y_i- f(X_i,Z_i; \theta)\|^2$ based on our formulation in Section \ref{problem_formulation}. More formally, the model seeks to maximize the observed-data log-likelihood: $\mathcal{L}(\theta)=\sum_{i=1}^n \log p_{\theta}(Y_i|X_i)$.
Specifically, let $p(Z)$ be the
prior distribution of $Z$. Let $p_\theta(Y|X,Z) \sim \mathcal{N}(f(X,Z; \theta), \sigma^2 I)$ be the conditional distribution of the noisy label $Y$ given $Z$ and $X$. The conditional distribution of $Y$ given $X$ is $p_\theta(Y|X) = \int p(Z) p_\theta(Y|X,Z) dZ$ with the latent variable $Z$ integrated out. 

The gradient of $\mathcal{L}(\theta)$ can be calculated according to the following identity: 
\begin{equation}
\label{update_phi}
\begin{aligned}
    \frac{\partial}{\partial \theta}\log p_{\theta}(Y|X)&= 
    \frac{1}{p_{\theta}(Y|X)}\frac{\partial}{\partial \theta}  p_{\theta}(Y|X)\\
    &=\text{E}_{p_{\theta}(Z|Y,X)} \left[\frac{\partial}{\partial \theta}\log p_{\theta}(Y,Z|X)\right].
\end{aligned}   
\end{equation}

The expectation term $\text{E}_{p_{\theta}(Z|Y,X)}$ is analytically intractable. The conventional way of training such a latent variable model is the variational inference, in which the intractable posterior distribution $p_{\theta}(Z|Y,X)$ is approximated by an extra trainable tractable neural network $p_{\phi}(Z|Y,X)$.
In this paper, we resort to
Monte Carlo average through drawing samples from the posterior distribution $p_{\theta}(Z|Y,X)$. This step corresponds to inferring the latent vector $Z$ of the generator for each training example. Specifically, we use Langevin Dynamics \cite{Neal2010MCMCUH} (a gradient-based Monte Carlo method) to sample $Z$. The Langevin Dynamics for sampling $Z\sim p_{\theta}(Z|Y,X)$ iterates:
\begin{equation}
\begin{aligned}
    Z_{t+1}=Z_{t}+ \frac{s^2}{2}\left[ \frac{\partial}{\partial Z}\log p_{\theta}(Y,Z_{t}|X)\right]+s \mathcal{N}(0,I_d),
    \label{langevin_dynamics}
\end{aligned}
\end{equation}
with 
\begin{equation}
\frac{\partial}{\partial Z}\log p_{\theta}(Y,Z|X) = \frac{1}{\sigma^2}(Y-f(X,Z;\theta))\frac{\partial}{\partial Z}f(X,Z) - Z,
\end{equation}
where $t$ and $s$ are the time step and step size of the Langevin Dynamics respectively. In each training iteration, for a given data pair $(X_i,Y_i)$, we run $l$ steps of Langevin Dynamics to infer $Z_i$. The Langevin Dynamics is initialized with Gaussian white noise (\ie, cold start) or the result of $Z_i$ obtained from the previous iteration (\ie, warm start). With the inferred $Z_i$ along with $(X_i,Y_i)$, the gradient used to update the model parameters $\theta$ is:
\begin{equation} 
\label{eq:learn}
\begin{aligned}
\frac{\partial}{\partial \theta} \mathcal{L}(\theta)& \approx \sum_{i=1}^{n}  \frac{\partial}{\partial \theta} \log p_\theta(Y_i,X_i|Z_i),  \\
&= \sum_{i=1}^{n} \frac{1}{\sigma^2}(Y_i-f(X_i,Z_i;\theta))\frac{\partial}{\partial \theta}f(X_i,Z_i).
\end{aligned}
\end{equation}

\begin{algorithm}
\small
\caption{Alternating back-propagation for noise-aware encoder-decoder}
\label{alg:proposed_algorithm}
\textbf{Input}: Dataset with noisy labels $D = \{(X_i,Y_i)\}_{i=1}^n$, learning epochs $K$, number of Langevin steps $l$, Langevin step size $s$, learning rate $\gamma$

\textbf{Output}: Network parameters $\theta=\{\theta_1,\theta_2\}$, and the inferred latent vectors $\{Z_i\}_{i=1}^n$

\begin{algorithmic}[1]
\State Initialize $\theta_1$ with the VGG16-Net\cite{VGG} for image classification, $\theta_2$ with a truncated Gaussian distribution, and $Z_i$ with a standard Gaussian distribution.
\For{$k = 1,...,K$}
\State \textbf{Inferential back-propagation}: For each $i$, run $l$ steps of Langevin Dynamics with a step size $s$ to sample $Z_i \sim p_{\theta}(Z_i|Y_i,X_i)$ following Eq. (\ref{langevin_dynamics}), with $Z_i$ initialized as a Gaussian white noise or the result from previous iteration.
\State \textbf{Learning back-propagation}: Update model parameters $\theta$ by Adam \cite{kingma2014adam} optimizer with a learning rate $\gamma$ and  the gradient $\frac{\partial}{\partial \theta} [\mathcal{L}(\theta) - \lambda l_{s}(X,S;\theta)]$, where the gradient of $\mathcal{L}(\theta)$ is computed according to Eq. (\ref{eq:learn}).
\EndFor
\end{algorithmic}
\end{algorithm}

To encourage the latent output $S$ of the encoder-decoder $f_1$ to be a meaningful saliency map,
we add a negative edge-aware smoothness loss \cite{occlusion_aware} defined on $S$ to the log-likelihood objective $\mathcal{L}(\theta)$. The smoothness loss serves as a regularization term to avoid a trivial decomposition of $S$ and $\Delta$ given $Y$.
Following \cite{occlusion_aware}, we use first-order derivatives (\ie, edge information) of both the latent clean saliency map $S$ and the input image $X$ to compute the smoothness loss 
\begin{equation}
\label{smoothness_loss}
    l_s(X, S) = \sum_{u,v} \sum_{d\in{x,y}} \Psi(|\partial_d S_{u,v}|e^{-\alpha |\partial_d X_{u,v}|}),
\end{equation}
\noindent
where $\Psi$ is the Charbonnier penalty formula, defined as $\Psi(s) = \sqrt{s^2+1e^{-6}}$, $(u,v)$ represent pixel coordinates, 
and $d$ indexes over the partial derivative in $x$ and $y$ directions. We estimate $\theta$ by gradient ascent on $\mathcal{L}(\theta) - \lambda l_{s}(X,S;\theta)$.  
In practice, we set $\lambda = 0.7$, and $\alpha=10$ in Eq. (\ref{smoothness_loss}).

The whole process of updating both $\{Z_i\}$ and $\theta=\{\theta_1,\theta_2\}$ is summarized in Algorithm \ref{alg:proposed_algorithm}, which is implemented as alternating back-propagation, because both gradients in Eq. (\ref{langevin_dynamics}) and (\ref{eq:learn}) can be computed via back-propagation.

\subsection{Comparison with Variational Inference}
\label{vae_comparison}
The proposed model can also be learned in a variational inference framework, where the intractable $p_{\theta}(Z|Y,X)$ in Eq. \ref{update_phi} is approximated by a tractable $q_{\phi}(Z|Y,X)$, such as $q_{\phi}(Z|Y,X) \sim \mathcal{N}(\mu_{\phi}(Y,X), \text{diag} (v_{\phi}(Y,X)))$, where both $\mu_{\phi}$ and $v_{\phi}$ are bottom-up networks that map $(X,Y)$ to $Z$, with $\phi$ standing for all parameters of the bottom-up networks. The objective of  variational inference is:
\begin{equation}
\begin{aligned} 
&\min_{\theta} \min_{\phi} \text{KL}(q_{\text{data}}(Y|X)p_{\phi}(Z|Y,X)\Vert p_{\theta}(Z, Y|X)) =\\
&\min_{\theta} \min_{\phi} \text{KL}(q_{\text{data}}(Y|X) \Vert p_{\theta}(Y|X)) + \text{KL}( p_{\phi}(Z|Y,X) \Vert p_{\theta}(Z|Y,X) ).
\end{aligned} 
\label{eq:VAE}
\end{equation}

 Recall that the maximum likelihood learning in our algorithm is equivalent to minimizing $\text{KL}(q_{\text{data}}(Y|X)\Vert p_{\theta}(Y|X))$, where $q_{\text{data}}(Y|X)$ is the conditional training data distribution. The accuracy of variational inference in Eq. \ref{eq:VAE} depends on the accuracy of an approximation of the true posterior distribution $p_{\theta}(Z|Y,X)$ by the inference model $p_{\phi}(Z|Y,X)$. Theoretically, the variational inference is equivalent to the maximum likelihood solution, when $\text{KL}( p_{\phi}(Z|Y,X) \Vert p_{\theta}(Z|Y,X))=0$. However, in practice, there is always a gap between them due to the design of the inference model and the optimization difficulty. Therefore, without relying on an extra assisting model, our alternating back-propagation algorithm is more natural, straightforward and computationally efficient than variational inference. We refer readers to \cite{xie2020representation} for a comprehensive tutorial on latent variable models.

\begin{figure*}[!htp]
   \begin{center}
   {\includegraphics[width=0.87\linewidth]{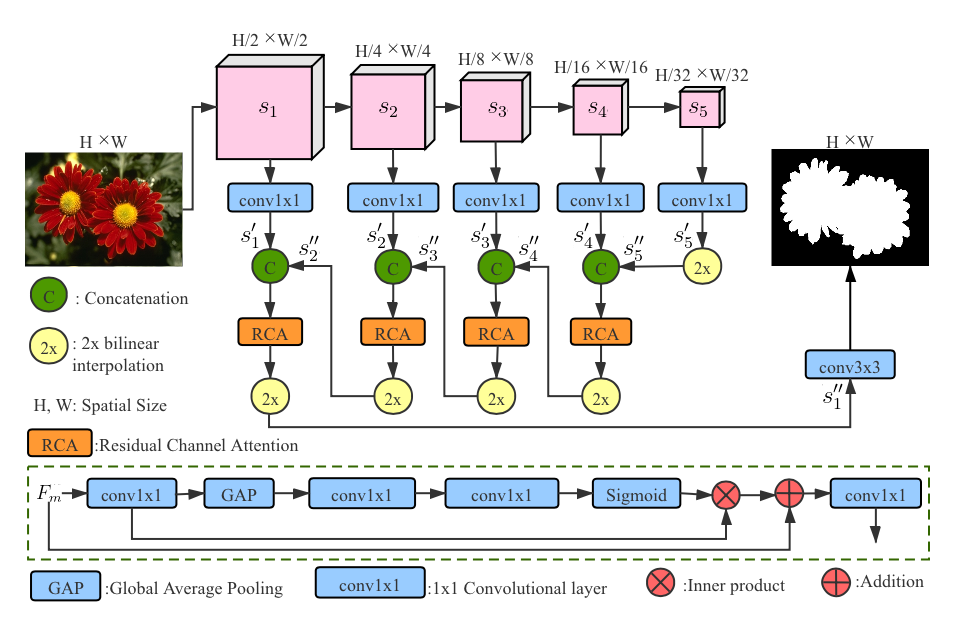}}
   \end{center}
   \caption{An illustration of the encoder-decoder-based saliency detection network (Green part in Fig.\ref{fig:illustration}).
   } 
   \label{fig:network_overview}
\end{figure*}
\subsection{Network Architectural Design}
\label{encoder_decoder_model}
We now introduce the architectural designs of the encoder-decoder network ($f_1$ in Eq. \ref{eq:1}, or the green encoder-decoder in Fig. \ref{fig:illustration}) and the noise generator network ($f_2$ in Eq. \ref{eq:2}, or the yellow decoder in Fig. \ref{fig:illustration}) in this section.

\textbf{Noise Generator:}
We construct the noise generator by using four cascaded deconvolutional layers, with a tanh activation function at the end to generate a noise map $\Delta$ in the range of $[-1,1]$. Batch normalization and ReLU layers are added between two nearby deconvolutional layers. The dimensionality of the latent variable $d=8$.

\textbf{Encoder-Decoder Network:} Most existing deep saliency prediction networks are based on widely used backbone networks, including the VGG16-Net \cite{VGG}, ResNet \cite{ResHe2015}, etc. Due to stride operations and multiple pooling layers used in these deep architectures, the 
saliency maps that are generated directly using the above backbone networks are low in spatial resolution, causing blurred edges.
To overcome this, we propose an encoder-decoder-based framework with the VGG16-Net \cite{VGG} as the backbone as shown in Fig.~\ref{fig:network_overview}. We denote the last convolutional layer of each convolutional group
of VGG16-Net by $s_1,s_2,...,s_5$ (corresponding to \enquote{relu1\_2}, \enquote{relu2\_2}, \enquote{relu3\_3}, \enquote{relu4\_3}, and \enquote{relu5\_3}, respectively). To reduce the channel dimension of $s_m$, a $1\times1$ convolutional layer is used to transform $s_m$ to $s'_m$ of channel dimension $32$.
Then a Residual Channel Attention (RCA) module \cite{RCAN_eccv} is adopted to effectively fuse the intermediate high- and low-level features. Specifically, given the high- and low-level feature maps $s'_{m}$ and $s'_{m-1}$, we first upsample $s'_{m}$ to $s''_{m}$, which has the same spatial resolution as $s'_{m-1}$, by bilinear interpolation.
Then we concatenate $s''_{m}$ and $s'_{m-1}$ to form a new feature map $F_m$.
Similar to \cite{RCAN_eccv}, we feed $F_m$ to the RCA block
to achieve the discriminative feature extraction. Inside each channel attention block, we perform \enquote{squeeze and excitation} \cite{hu2018senet} by first \enquote{squeezing} the input feature map $F_m$ to be half of the original channel size 
to obtain better nonlinear interactions across channels, and then \enquote{exciting} the squeezed feature map back to the original channel size.
By adding a $3\times3$ convolutional layer to the lowest level of the RCA module,
we obtain a one-channel saliency map $S_i=f_1(X_i;\theta_1)$.


\section{Experiments}

\subsection{Experimental Setup}
\textbf{Datasets:} We evaluate our performance on five saliency benchmark datasets. We use 10,553 images from the DUTS dataset~\cite{imagesaliency} for training, and we generate noisy labels from images using handcrafted feature based-methods, such as RBD \cite{Background-Detection:CVPR-2014}, MR \cite{Manifold-Ranking:CVPR-2013} and GS \cite{GS_Sal}
due to their high efficiencies. Testing datasets include the DUTS testing set, ECSSD \cite{Hierarchical:CVPR-2013}, DUT \cite{Manifold-Ranking:CVPR-2013}, HKU-IS \cite{MDF:CVPR-2015}
and THUR \cite{THUR}.

\textbf{Evaluation Metrics:} Four metrics are used to evaluate the performance of our method and the competing methods, including two widely used metrics, \ie, Mean Absolute Error ($\mathcal{M}$) and mean F-measure ($F_{\beta}$), and two newly released structure-aware metrics: mean E-measure ($E_{\xi}$) \cite{Fan2018Enhanced} and S-measure ($S_{\alpha}$) \cite{fan2017structure}.

\textbf{Training Details:}
Each input image is rescaled to $352\times 352$ pixels. The encoder part in Fig. \ref{fig:network_overview} is
initialized
using the VGG16-Net weights pretrained for image classification \cite{VGG}. The weights of other layers are initialized using the \enquote{truncated Gaussian} policy, and the biases are initialized to be zeros. We use the Adam \cite{kingma2014adam} optimizer with  a momentum equal to 0.9, and decrease the learning rate $\gamma$ by 10\% after running 80\% of the maximum epochs $K=20$. The learning rate is initialized to be 0.0001. The number of Langevin steps $l$ is 6. The Langevin step size $s$ is 0.3. The $\sigma$ in Eq.(\ref{eq:3}) is 0.1. The whole training takes 8 hours with a batch size 10 on a PC with an NVIDIA GeForce RTX GPU. We use the PaddlePaddle \cite{paddle} deep learning platform.

\begin{table*}[t!]
  \centering
  \scriptsize
  \renewcommand{\arraystretch}{1.2}
  \renewcommand{\tabcolsep}{0.4mm}
  \caption{Benchmarking performance comparison.
  Bold numbers represent best performance methods.
  $\uparrow \& \downarrow$ denote larger and smaller is better, respectively.
  }
  \begin{tabular}{lr|ccccccc|cccccc}
  \toprule[1pt]
  &  &\multicolumn{7}{c|}{Fully Suppervised Models}&\multicolumn{6}{c}{Weakly Sup./Unsup. Models} \\
    & Metric &
   DGRL  &  NLDF &
   MSNet & CPD & AFNet & SCRN  & BASNet & C2S & WSI & WSS & MNL & MSW & Ours\\
  &  & \cite{Wang_2018_CVPR}        & \cite{NLDF}       & \cite{MSNet_Sal}          & \cite{CPD_Sal}              & \cite{AFNet_Sal} &
        \cite{SCRN_iccv}   & \cite{BASNet_Sal}                 & \cite{xin2018c2s} 
        & \cite{Guanbin_weaksalAAAI}   & \cite{imagesaliency} &\cite{Zhang_2018_CVPR} & \cite{MSW_Sal} & \\
  \hline
  \multirow{4}{*}{\begin{sideways}\textit{DUTS}\end{sideways}}
    & $S_{\alpha}\uparrow$    & .8460 & .8162 & .8617 & .8668 & .8671 & \textbf{.8848}  & .8657 & .8049 & .6966 & .7484& .8128 & .7588 & \textbf{.8276} \\
    & $F_{\beta}\uparrow$     & .7898 & .7567 & .7917 & .8246 & .8123 & \textbf{.8333} & .8226 & .7182 & .5687 & .6330& .7249 & .6479& \textbf{.7467}   \\
    & $E_{\xi}\uparrow$       & .8873 & .8511 & .8829 & \textbf{.9021} & .8928 & .8996 & .8955 & .8446 & .6900 & .8061 & .8525 & .7419& \textbf{.8592} \\
    & $\mathcal{M}\downarrow$ & .0512 & .0652 & .0490 & .0428 & .0457 & \textbf{.0398} & .0476 & .0713 & .1156 & .1000 & .0749 & .0912 & \textbf{.0601} \\ \hline
  \multirow{4}{*}{\begin{sideways}\textit{ECSSD}\end{sideways}}
    & $S_{\alpha}\uparrow$    & .9019 & .8697 & .9048 & .9046 & .9074 & \textbf{.9204} & .9104 & - & .8049 & .8081& .8456 & .8246 & \textbf{.8603} \\
    & $F_{\beta}\uparrow$     & .8978 & .8714 & .8856 & .9076 & .9008 & .9103 & \textbf{.9128} & - & .7621 & .7744& .8098 & .7606& \textbf{.8519}   \\
    & $E_{\xi}\uparrow$       & .9336 & .8955 & .9218 & .9321 & .9294 & .9333 & \textbf{.9378} & - & .7921 & .8008& .8357 & .7876& \textbf{.8834}   \\
    & $\mathcal{M}\downarrow$ & .0447 & .0655 & .0479 & .0434 & .0450 & .0407 & \textbf{.0399} & - & .1137 & .1055 & .0902 & .0980& \textbf{.0712}  \\ \hline
  \multirow{4}{*}{\begin{sideways}\textit{DUT}\end{sideways}}
    & $S_{\alpha}\uparrow$    & .8097 & .7704 & .8093 & .8177 & .8263 & \textbf{.8365} & .8362 & .7731 & .7591 & .7303& .7332 & .7558& \textbf{.7914}   \\
    & $F_{\beta}\uparrow$     & .7264 & .6825 & .7095 & .7385 & .7425 & .7491 & \textbf{.7668} & .6649 & .6408 & .5895 & .5966 & .5970& \textbf{.7007}  \\
    & $E_{\xi}\uparrow$       & .8446 & .7983 & .8306 & .8450 & .8456 & .8474 & \textbf{.8649} & .8100 & .7605 & .7292 & .7124 & .7283& \textbf{.8158}  \\
    & $\mathcal{M}\downarrow$ & .0632 & .0796 & .0636 & .0567 & .0574 & \textbf{.0560} & .0565 & .0818 & .0999 & .1102 & .1028 & .1087& \textbf{.0703} \\ \hline
  \multirow{4}{*}{\begin{sideways}\textit{HKU-IS}\end{sideways}}
    & $S_{\alpha}\uparrow$    & .8968 & .8787 & .9065 & .9039 & .9053 & \textbf{.9158} & .9089 & .8690 & .8079 & .8223& .8602 & .8182& \textbf{.8901}  \\
    & $F_{\beta}\uparrow$     & .8844 & .8711 & .8780 & .8948 & .8877 & .8942 & \textbf{.9025} & .8365 & .7625 & .7734 & .8196 & .7337& \textbf{.8782}  \\
    & $E_{\xi}\uparrow$       & .9388 & .9139 & .9304 & .9402 & .9344 & .9351 & \textbf{.9432} & .9103 & .7995 & .8185 & .8579 & .7862 & \textbf{.9191} \\
    & $\mathcal{M}\downarrow$ & .0374 & .0477 & .0387 & .0333 & .0358 & .0337 & \textbf{.0322} & .0527 & .0885 & .0787 & .0650 & .0843 & \textbf{.0428} \\ \hline
   \multirow{4}{*}{\begin{sideways}\textit{THUR}\end{sideways}}
    & $S_{\alpha}\uparrow$    & .8162 & .8008 & .8188 & .8311 & .8251 & \textbf{.8445} & .8232  & .7922 & - & .7751 & .8041 & -& \textbf{.8101}  \\
    & $F_{\beta}\uparrow$     & .7271 & .7111 & .7177 & .7498 & .7327 & \textbf{.7584} & .7366 & .6834 & - & .6526 & 6911 & -& \textbf{.7187}  \\
    & $E_{\xi}\uparrow$       & .8378 & .8266 & .8288 & .8514 & .8398 & \textbf{.8575} & .8408 & .8107 & - & .7747 & .8073 & - & \textbf{.8378} \\
    & $\mathcal{M}\downarrow$ & .0774 & .0805 & .0794 & \textbf{.0635} & .0724 & .0663 & .0734  & .0890 & - & .0966 & .0860 & - & \textbf{.0703} \\
    \bottomrule[1pt]
  \end{tabular}
  \label{tab:deep_unsuper_Performance_Comparison}
\end{table*}

\subsection{Comparison with the State-of-the-art Methods}
\begin{figure*}[!htp]
   \begin{center}
   {\includegraphics[width=0.243\linewidth]{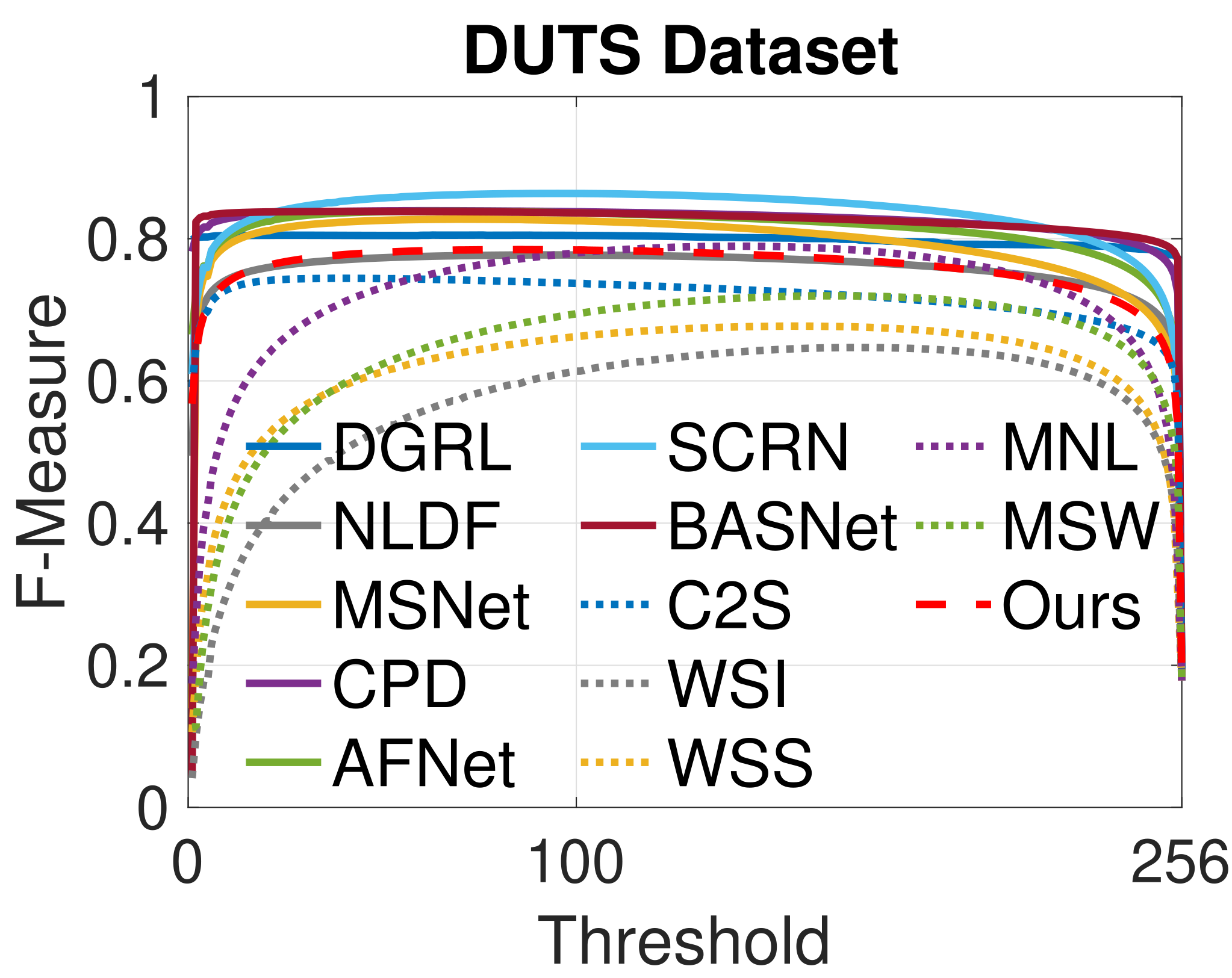}}
   {\includegraphics[width=0.243\linewidth]{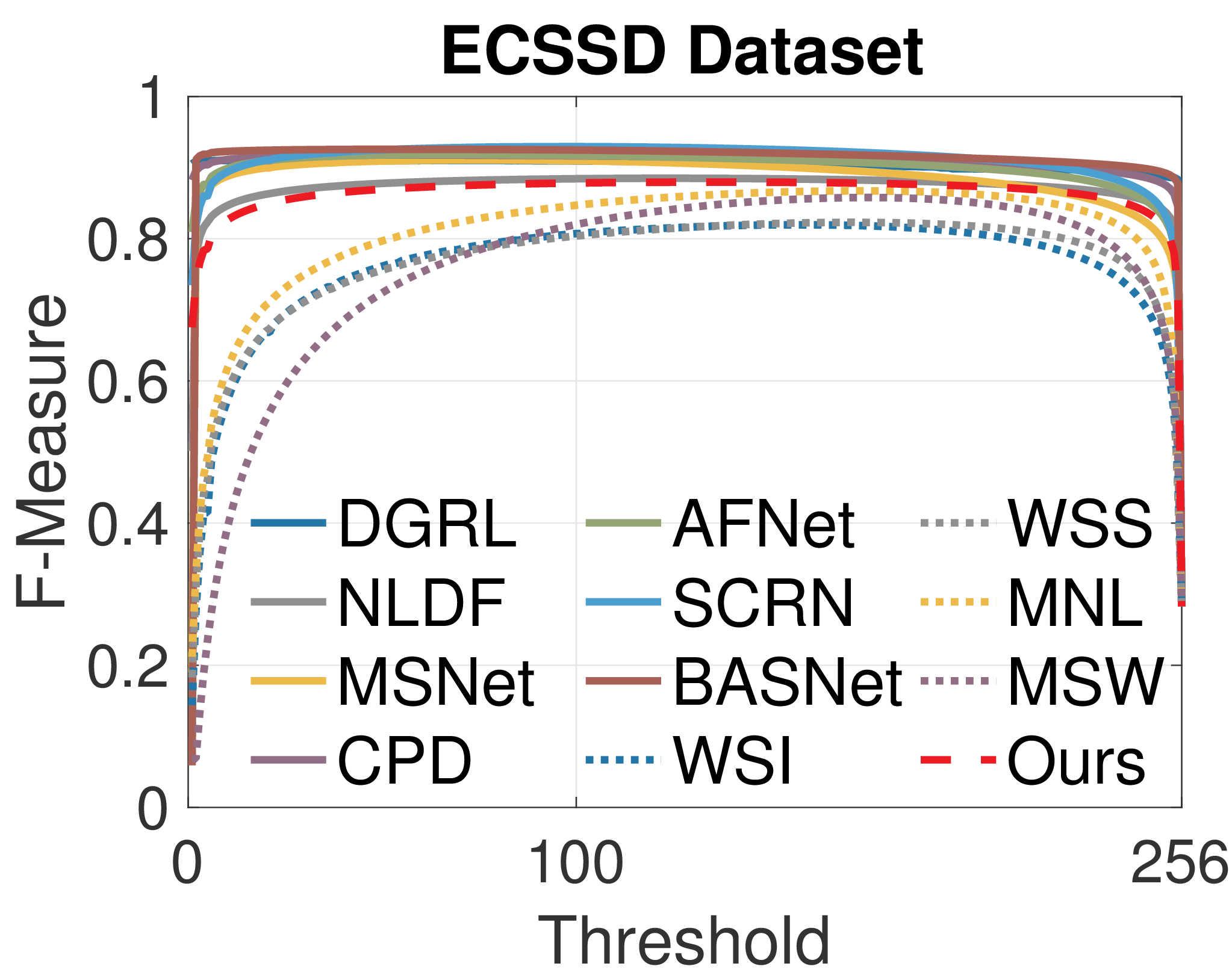}}
   {\includegraphics[width=0.243\linewidth]{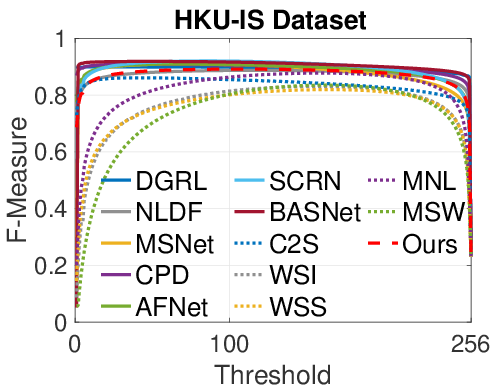}}
   {\includegraphics[width=0.243\linewidth]{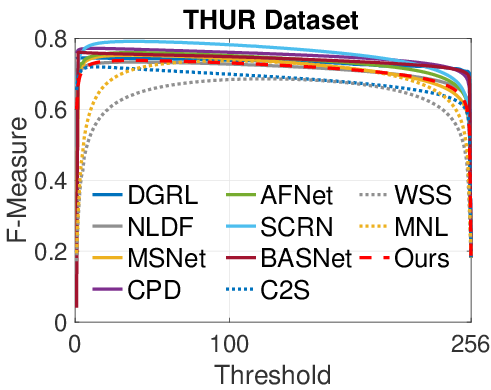}}
   {\includegraphics[width=0.243\linewidth]{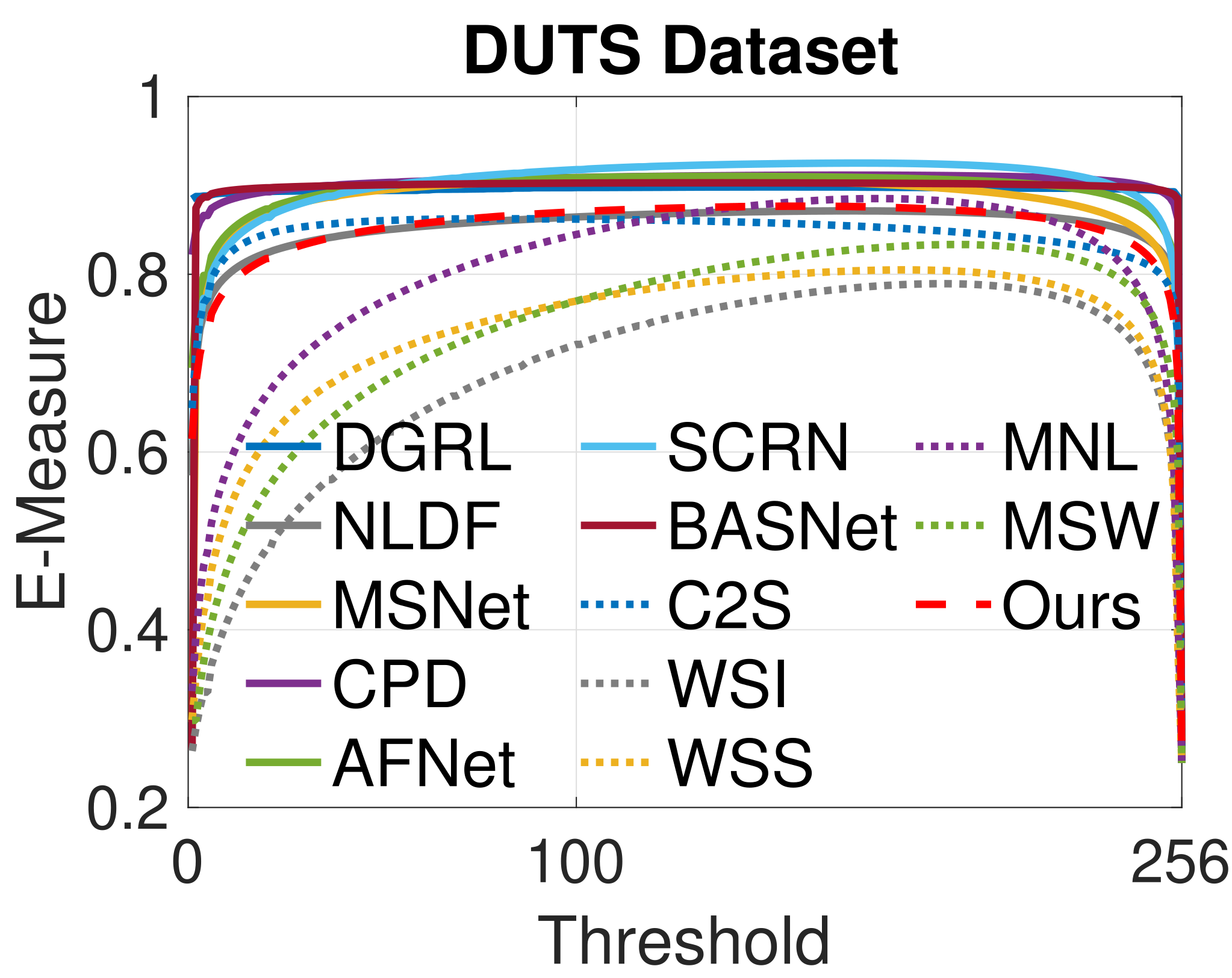}}
   {\includegraphics[width=0.243\linewidth]{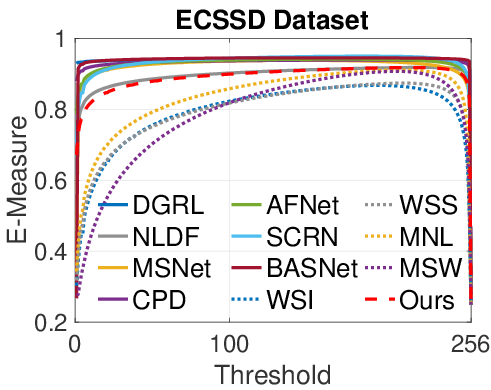}}
   {\includegraphics[width=0.243\linewidth]{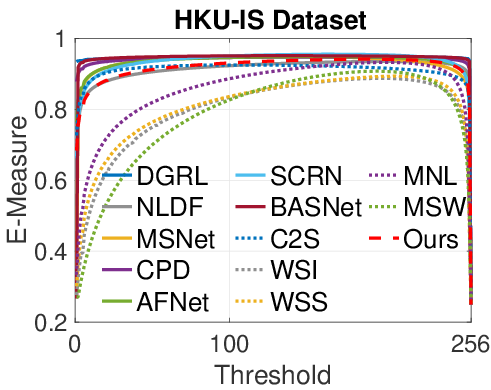}}
   {\includegraphics[width=0.243\linewidth]{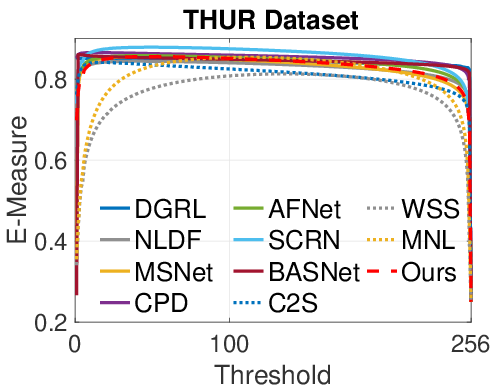}}
   \end{center}
   \caption{F-measure and E-measure curves on four datasets (DUTS, ECSSD, HKU-IS, THUR).
   Best viewed in color on screen. 
   }
   \label{fig:ef_curve}
\end{figure*}
We compare our method with seven fully supervised deep saliency prediction models and five weakly supervised/unsupervised saliency prediction models, and their performances are shown in Table \ref{tab:deep_unsuper_Performance_Comparison} and Fig. \ref{fig:ef_curve}. Table \ref{tab:deep_unsuper_Performance_Comparison} shows that compared with the weakly supervised/unsupervised models, the proposed method achieves the best performance, especially on DUTS and HKU-IS datasets, where our method achieves an approximately 2\% performance improvement for S-measure, and a 4\% improvement for mean F-measure. Further, the proposed method even achieves comparable performances with some newly released fully supervised models. For example, we achieve comparable performance with NLDF \cite{NLDF} and DGRL \cite{Wang_2018_CVPR} on all the five benchmark datasets.
Fig.\ref{fig:ef_curve} shows the 256-dimensional F-measure and E-measure (where the x-axis represents threshold for saliency map binarization) of our method and the competing methods on four datasets, where the weakly supervised/unsupervised methods are represented by dotted curves. We can observe that the performances of the fully supervised models are better than those of the weakly supervised/unsupervised models. As shown in Fig.\ref{fig:ef_curve}, our performance shows stability with different thresholds relative to the existing methods, indicating the robustness of our model.

\begin{figure*}[!t]
\tabcolsep5pt
\arrayrulewidth1pt
  \begin{center}
  \begin{tabular}{c@{ }: c@{ } c@{ } c@{ }  c@{ }: c@{ } c@{ }  c@{ } :c@{ } c@{ }}
  \multicolumn{1}{c}{} &  \multicolumn{4}{c}{Fully Supervised} & \multicolumn{3}{c}{Weak/Un Supervised} & &  \\
  {\includegraphics[width=0.085\linewidth]{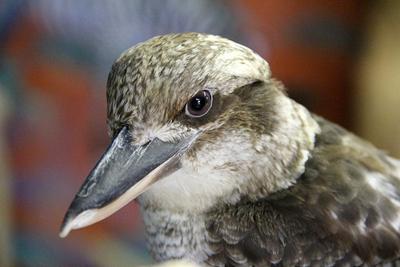}}&
  {\includegraphics[width=0.085\linewidth]{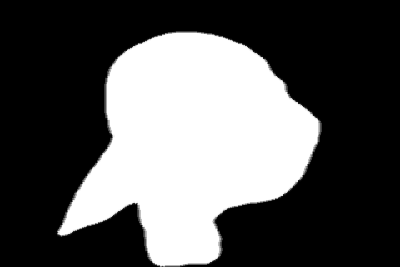}}&
  {\includegraphics[width=0.085\linewidth]{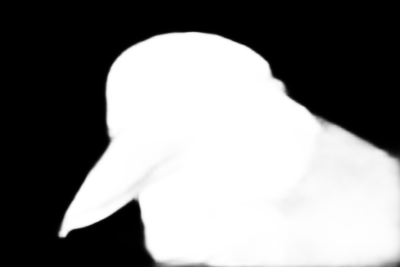}}&
  {\includegraphics[width=0.085\linewidth]{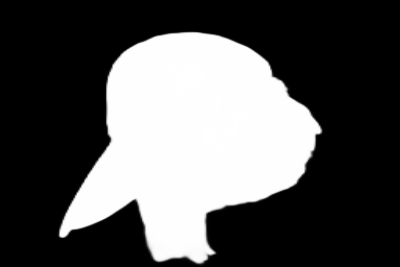}}&
  {\includegraphics[width=0.085\linewidth]{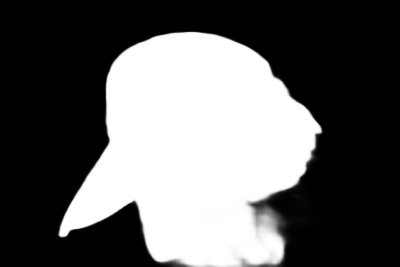}}&
  {\includegraphics[width=0.085\linewidth]{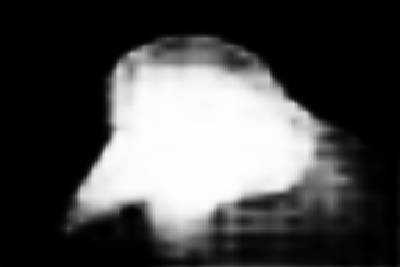}}&
  {\includegraphics[width=0.085\linewidth]{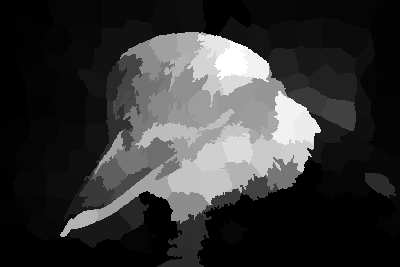}}&
  {\includegraphics[width=0.085\linewidth]{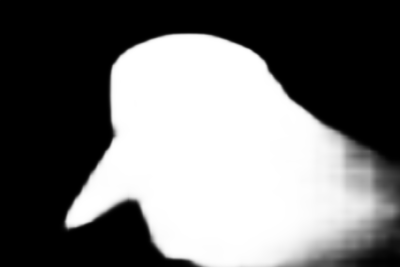}}&
  {\includegraphics[width=0.085\linewidth]{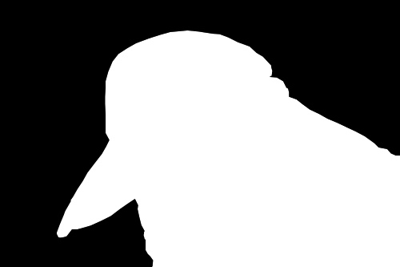}}&
  {\includegraphics[width=0.085\linewidth]{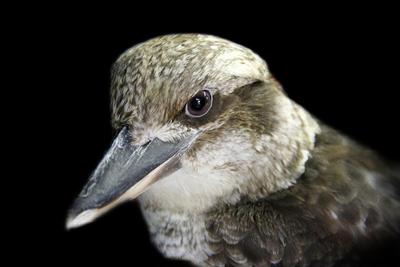}}\\
  {\includegraphics[width=0.085\linewidth]{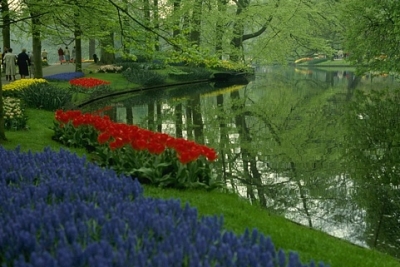}}&
  {\includegraphics[width=0.085\linewidth]{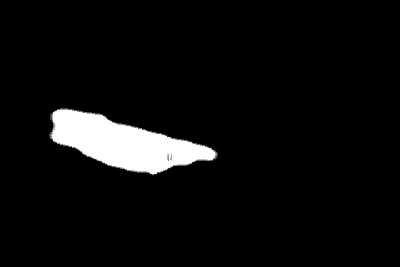}}&
  {\includegraphics[width=0.085\linewidth]{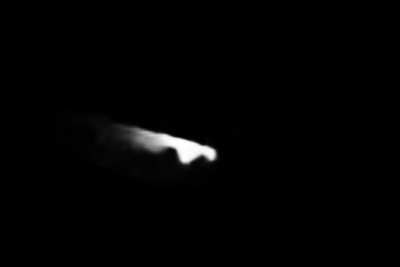}}&
  {\includegraphics[width=0.085\linewidth]{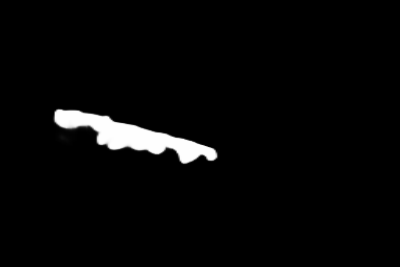}}&
  {\includegraphics[width=0.085\linewidth]{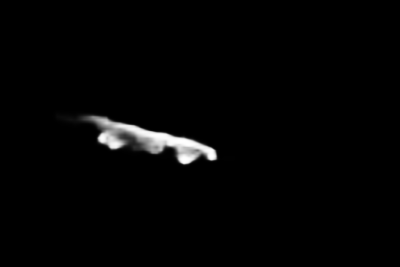}}&
  {\includegraphics[width=0.085\linewidth]{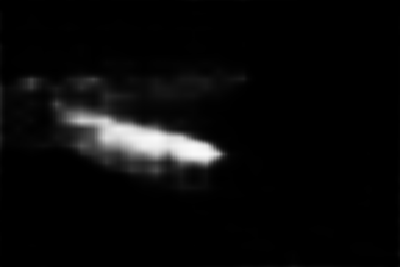}}&
  {\includegraphics[width=0.085\linewidth]{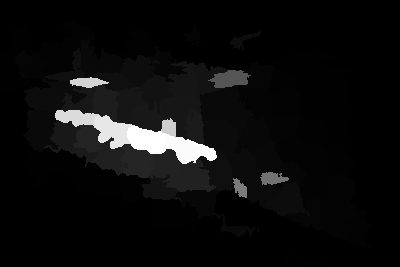}}&
  {\includegraphics[width=0.085\linewidth]{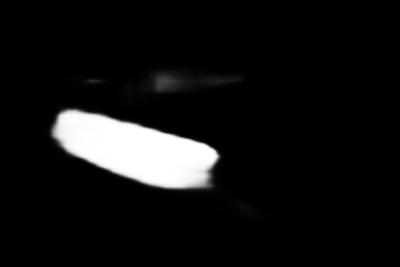}}&
  {\includegraphics[width=0.085\linewidth]{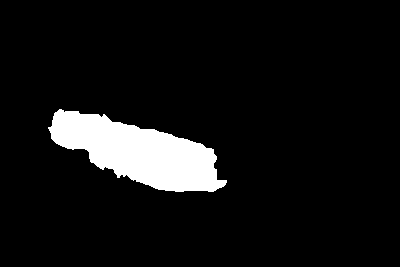}}&
  {\includegraphics[width=0.085\linewidth]{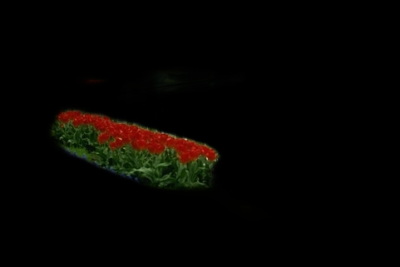}}\\
  {\includegraphics[width=0.085\linewidth]{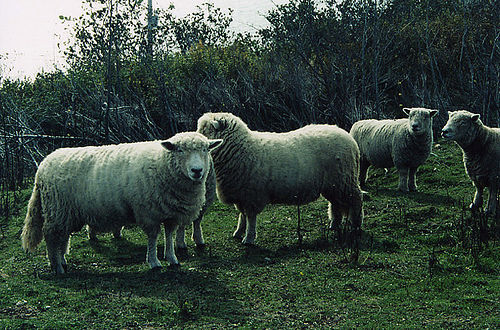}}&
  {\includegraphics[width=0.085\linewidth]{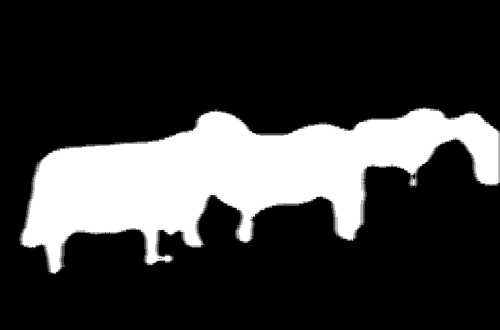}}&
  {\includegraphics[width=0.085\linewidth]{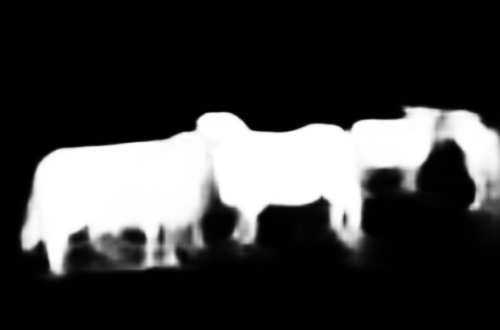}}&
  {\includegraphics[width=0.085\linewidth]{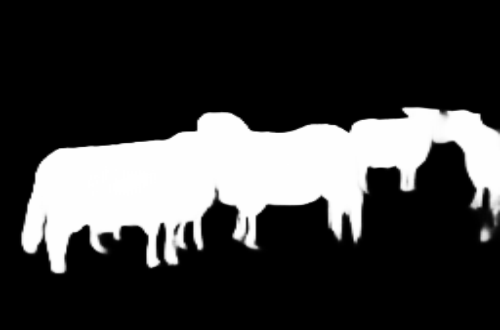}}&
  {\includegraphics[width=0.085\linewidth]{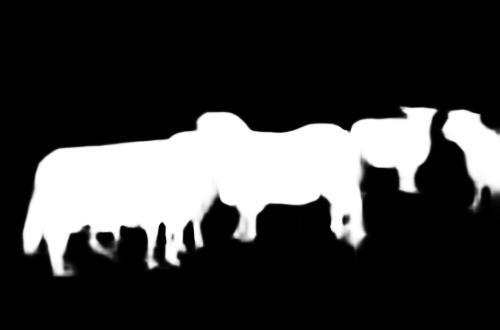}}&
  {\includegraphics[width=0.085\linewidth]{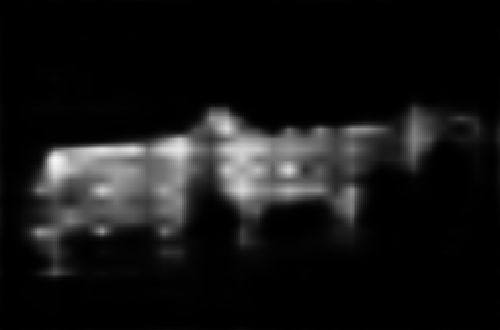}}&
  {\includegraphics[width=0.085\linewidth]{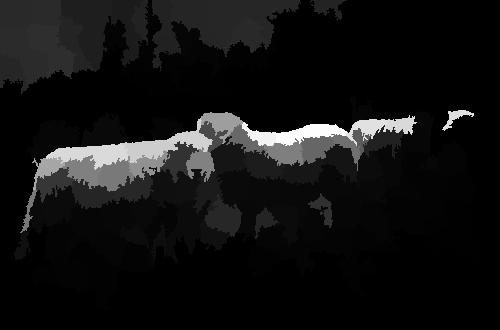}}&
  {\includegraphics[width=0.085\linewidth]{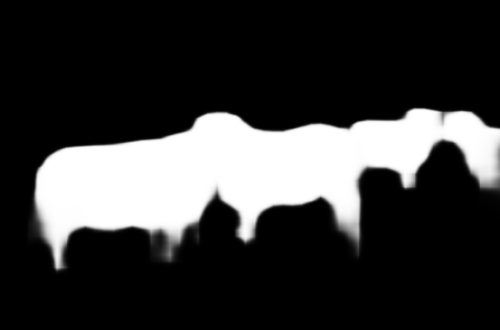}}&
  {\includegraphics[width=0.085\linewidth]{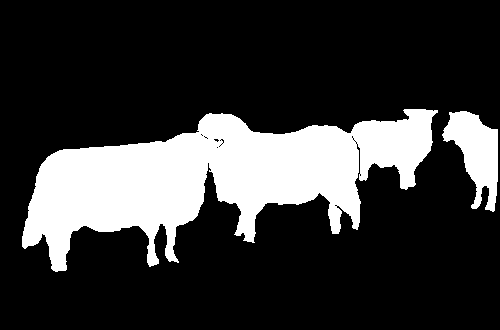}}&
  {\includegraphics[width=0.085\linewidth]{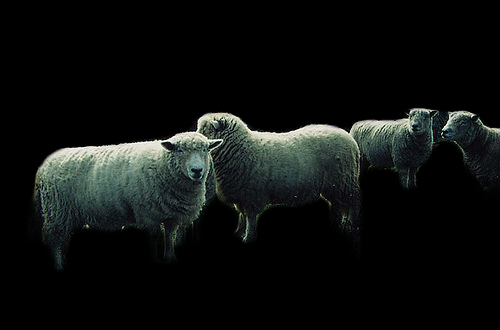}}\\
  {\includegraphics[width=0.085\linewidth]{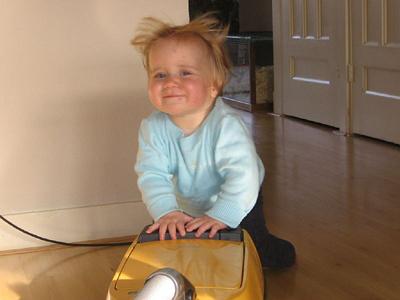}}&
  {\includegraphics[width=0.085\linewidth]{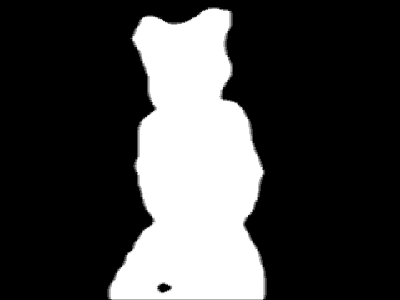}}&
  {\includegraphics[width=0.085\linewidth]{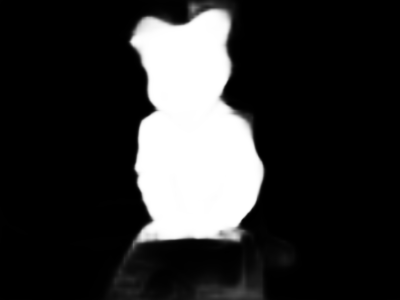}}&
  {\includegraphics[width=0.085\linewidth]{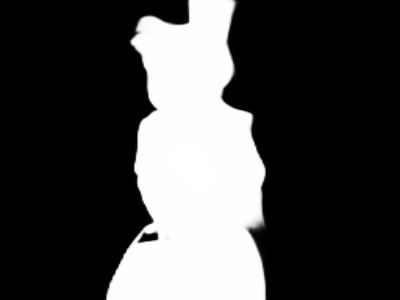}}&
  {\includegraphics[width=0.085\linewidth]{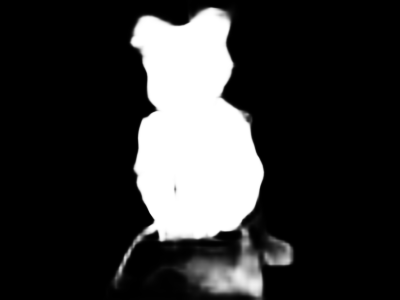}}&
  {\includegraphics[width=0.085\linewidth]{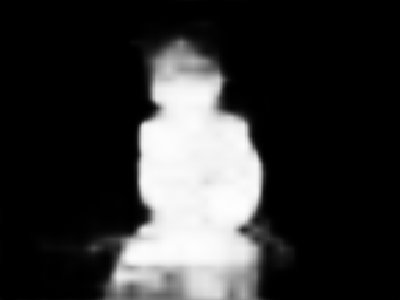}}&
  {\includegraphics[width=0.085\linewidth]{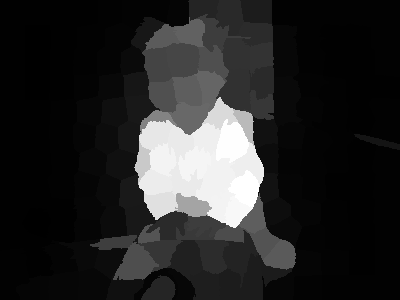}}&
  {\includegraphics[width=0.085\linewidth]{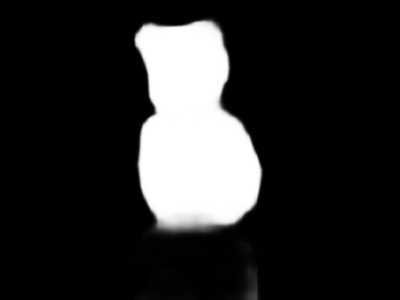}}&
  {\includegraphics[width=0.085\linewidth]{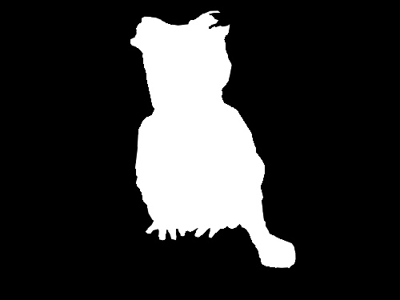}}&
  {\includegraphics[width=0.085\linewidth]{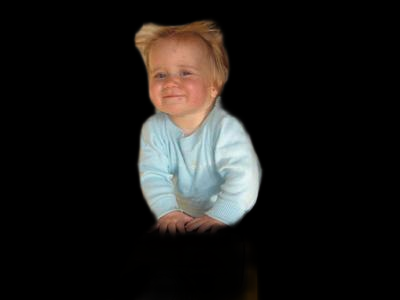}}\\
  {\includegraphics[width=0.085\linewidth]{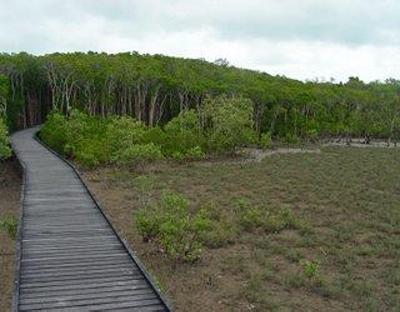}}&
  {\includegraphics[width=0.085\linewidth]{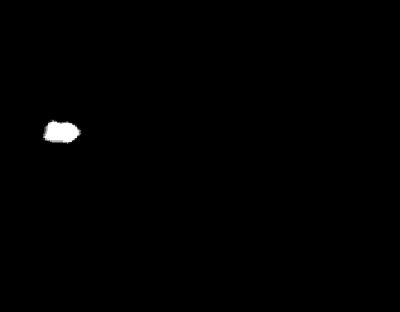}}&
  {\includegraphics[width=0.085\linewidth]{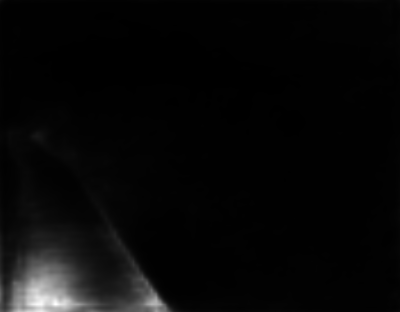}}&
  {\includegraphics[width=0.085\linewidth]{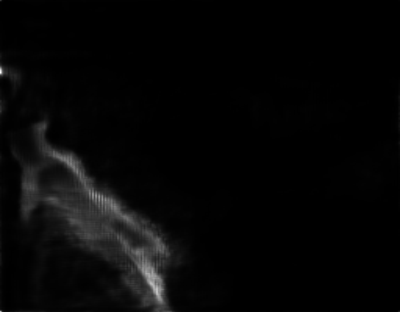}}&
  {\includegraphics[width=0.085\linewidth]{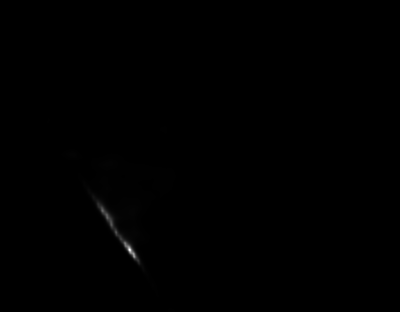}}&
  {\includegraphics[width=0.085\linewidth]{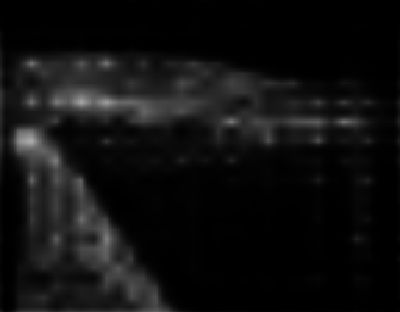}}&
  {\includegraphics[width=0.085\linewidth]{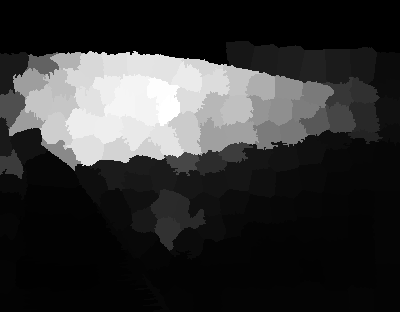}}&
  {\includegraphics[width=0.085\linewidth]{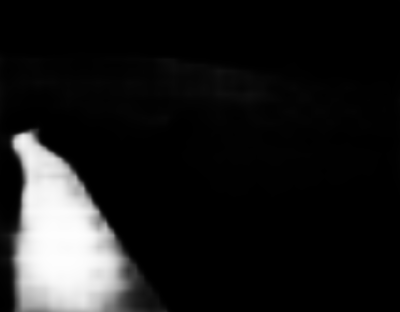}}&
  {\includegraphics[width=0.085\linewidth]{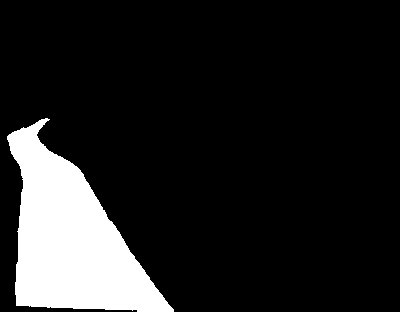}}&
  {\includegraphics[width=0.085\linewidth]{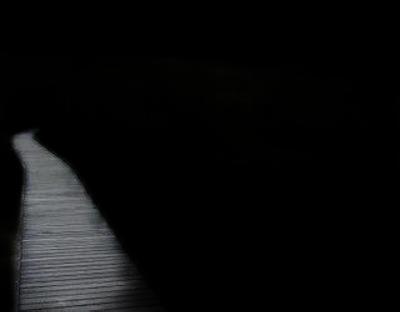}}\\
  \footnotesize{Image} & \footnotesize{DGRL} & \footnotesize{SCRN} & \footnotesize{BSNet} & \footnotesize{CPD} & \footnotesize{MSW} & \footnotesize{RBD} & \footnotesize{Ours} & \footnotesize{GT} & \footnotesize{Seg}\\
  \end{tabular}
  \end{center}
\caption{\small Comparison of saliency predictions, where each row displays an input image, its predicted saliency maps by four fully supervised competing methods (DGRL, SCRN, BASNet, and CPD), one weakly (MSW) and one unsupervised (RBD) methods, our prediction (Ours), the ground truth (GT) saliency map and our segmented foreground image (Seg).}
  \label{fig:saliency_compare}
\end{figure*}

Figure~\ref{fig:saliency_compare} demonstrates a qualitative comparison on several challenging cases. For example, the salient object in the first row is large, and connects to the image border. Most competing methods fail to segment the  border-connected region, while our method almost finds the whole salient region in this case. Also, salient object in the second row has a long and narrow shape, which is challenging to some competing methods. Our method performs very well and precisely detect the salient object.

\subsection{Ablation Study}
\label{sec:ablation_study}

We conduct the following experiments for an ablation study.

\begin{table*}[t!]
  \centering
  \scriptsize
  \renewcommand{\arraystretch}{1.4}
  \renewcommand{\tabcolsep}{0.35mm}
  \caption{Ablation study. Some certain key components of the model are removed and the learned model is evaluated for saliency prediction in terms of $S_{\alpha}$,  $F_{\beta}$, $E_{\xi}$, and $\mathcal{M}$. $\uparrow \& \downarrow$ denote larger and smaller is better, respectively.}
  \begin{tabular}{l|cccc|cccc|cccc|cccc|cccc}
  \toprule[1pt]
   &\multicolumn{4}{c|}{DUTS}&\multicolumn{4}{c|}{ECSSD}&\multicolumn{4}{c|}{DUT}&\multicolumn{4}{c|}{HKU-IS}&\multicolumn{4}{c}{THUR} \\
  &
   $S_{\alpha}$  &  $F_{\beta}$ & $E_{\xi}$  & $\mathcal{M}$ &
   $S_{\alpha}$  &  $F_{\beta}$ & $E_{\xi}$  & $\mathcal{M}$ &
   $S_{\alpha}$  &  $F_{\beta}$ & $E_{\xi}$  & $\mathcal{M}$ &
   $S_{\alpha}$  &  $F_{\beta}$ & $E_{\xi}$  & $\mathcal{M}$ &
   $S_{\alpha}$  &  $F_{\beta}$ & $E_{\xi}$  & $\mathcal{M}$ \\
  Model&
   $\uparrow$  &  $\uparrow$ & $\uparrow$  & $\downarrow$ &
   $\uparrow$  &  $\uparrow$ & $\uparrow$  & $\downarrow$ &
   $\uparrow$  &  $\uparrow$ & $\uparrow$  & $\downarrow$ &
  $\uparrow$  &  $\uparrow$ & $\uparrow$  & $\downarrow$ &
   $\uparrow$  &  $\uparrow$ & $\uparrow$  & $\downarrow$ \\
   
  \hline
  $f_1$ &.644  & .453 &.632 & .157 &.685 & .559 &.650  & .174 &.679  & .497 &.663  & .147 &.706  & .572 &.674  & .143 &.665  & .472 &.656  & .151 \\
  $f_1\&l_s$ &.668  & .519 &.699  & .125 &.727  & .675 &.743  & .138 &.685  & .537 &.720  & .121 &.743  & .681 &.775  & .107 &.687  & .547 &.727  & .121 \\
   $f\&l_c$ &.813  & .725 &.806  & .075 &.846  & .810 &.836  & .090 &.733  & .597 &.712  & .103 &.860  & .820 &.858  & .065 &.804  & .691 &.807  & .086 \\
  Full &.828  & .747 &.859  & .060 &.860  & .852 &.883  & .071 &.791  & .701 &.816  & .070 &.890  & .878 &.919  & .043 &.810  & .719 &.838  & .070 \\
   \bottomrule[1pt]
  \end{tabular}
  \label{tab:ablation_study}
\end{table*}

\textbf{(1) Encoder-decoder $f_1$ only:} To study the effect of the noise generator, we evaluate the performance of the encoder-decoder (as shown in Fig. \ref{fig:network_overview}) directly learned from the noisy labels, without noise modeling or smoothness loss.  
The performance is shown in Table \ref{tab:ablation_study} with a label ``$f_1$'', which is clearly worse than ours.
This result is also consistent with the conclusion that deep neural networks is not robust to noise \cite{zhang_iclr}.

\textbf{(2) Encoder-decoder $f_1$ + smoothness loss $l_s$:} As an extension of method \enquote{$f_1$}, one can add the smoothness loss in Eq. (\ref{smoothness_loss}) as a regularization to better use the image prior information. We show the performance with a label \enquote{$f_1$ \& $l_s$} in Table \ref{tab:ablation_study}. We observe a performance improvement compared with \enquote{$f_1$}, which indicates the usefulness of the edge-aware smoothness loss.


\textbf{(3) Noisy-aware encoder-decoder without edge-aware smoothness loss:}
To study the effect of the smoothness regularization, we try to remove the smoothness loss from our model. As a result, we find that it will lead to trivial solutions \ie, $S_i=\mathbf{0}_{H\times W}$ for all training images.
 
\textbf{(4) Alternative smoothness loss:}
 We also replace our smoothness loss $l_s$ by a cross-entropy loss $l_c(S,X)$ that is also defined on the first-order derivative of the saliency map $S$ and that of the image $X$. 
 The performance is shown in Table \ref{tab:ablation_study} as \enquote{$f$ \& $l_c$}, which is better than or comparable with the existing weakly supervised/unsupervised methods shown in Table \ref{tab:deep_unsuper_Performance_Comparison}.
By comparing the performance of \enquote{$f$ \& $l_c$} with that of the full model, we observe that the smoothness loss $l_s(S,X)$ in Eq. \ref{smoothness_loss} works better than the cross-entropy loss $l_c(S,X)$. The former puts a soft constraint on their boundaries, while the latter has a strong effect on forcing both boundaries of $S$ and $X$ to be the same. Although the saliency boundary 
are usually aligned with the image boundary, but they are not exactly the same. A soft and indirect penalty for edge dissimilarity seems to be more useful. 

\subsection{Model Analysis}
We further explore our proposed model in this section. 

\textbf{(1) Learn the model from saliency labels generated by fully supervised pre-trained models:} One way to use our method is treating it as a boosting strategy for the current fully-supervised models. To verify this, we first generate saliency maps by using a pre-trained fully-supervised saliency network, \eg, BASNet \cite{BASNet_Sal}. We treat the outputs as noisy labels, on which we train our model.  
The performances are shown in Table \ref{tab:model_analysis} as $f$-BAS.
By comparing the performances of $f$-BAS with those of BASNet in Table \ref{tab:deep_unsuper_Performance_Comparison},
we find that $f$-BAS is comparable with or better than BASNet, which means that our method can further refine the outputs of the state-of-the-art pre-trained fully-supervised models if their performances are still far from perfect.

\begin{table*}[t!]
  \centering
  \scriptsize
  \renewcommand{\arraystretch}{1.4}
  \renewcommand{\tabcolsep}{0.30mm}
  \caption{Experimental results for model analysis. $\uparrow \& \downarrow$ denote larger and smaller is better, respectively.
  }
  \begin{tabular}{l|cccc|cccc|cccc|cccc|cccc}
  \hline
  \toprule[1pt]
   &\multicolumn{4}{c|}{DUTS}&\multicolumn{4}{c|}{ECSSD}&\multicolumn{4}{c|}{DUT}&\multicolumn{4}{c|}{HKU-IS}&\multicolumn{4}{c}{THUR} \\
  &
   $S_{\alpha}$  &  $F_{\beta}$ & $E_{\xi}$  & $\mathcal{M}$ &
   $S_{\alpha}$  &  $F_{\beta}$ & $E_{\xi}$  & $\mathcal{M}$ &
   $S_{\alpha}$  &  $F_{\beta}$ & $E_{\xi}$  & $\mathcal{M}$ &
   $S_{\alpha}$  &  $F_{\beta}$ & $E_{\xi}$  & $\mathcal{M}$ &
   $S_{\alpha}$  &  $F_{\beta}$ & $E_{\xi}$  & $\mathcal{M}$ \\
   Model&
   $\uparrow$  &  $\uparrow$ & $\uparrow$  & $\downarrow$ &
   $\uparrow$  &  $\uparrow$ & $\uparrow$  & $\downarrow$ &
   $\uparrow$  &  $\uparrow$ & $\uparrow$  & $\downarrow$ &
  $\uparrow$  &  $\uparrow$ & $\uparrow$  & $\downarrow$ &
   $\uparrow$  &  $\uparrow$ & $\uparrow$  & $\downarrow$ \\
  \hline
  $f$-BAS &.870  & .823 &.894 & .042 &.910 & .910 &.935  & .040 &.839  & .769 &.866  & .056 &.904  & .900 &.945  & .032 &.821  & .737 &.840  & .073 \\
  $f$-RBD &.824  & .753 &.854  & .066 &.869  & .856 &.890  & .070 &.776  & .675 &.799  & .082 &.886  & .863 &.918  & .047 &.803  & .700 &.823  & .082 \\
  $f$-MR &.814  & .759 &.839  & .064 &.857  & .856 &.876  & .073 &.762  & .669 &.779  & .079 &.972  & .866 &.901  & .050 &.794  & .696 &.804  & .086 \\
  $f$-GS &.787  & .740 &.811  & .071 &.826  & .836 &.843  & .087 &.737  & .652 &.753  & .083 &.837  & .843 &.865  & .062 &.804  & .723 &.840  & .071 \\
  RBD &.644  & .453 &.632  & .157 &.685  & .559 &.650  & .174 &.679  & .497 &.663  & .147 &.706  & .572 &.674  & .143 &.665  & .472 &.656  & .151 \\
  MR &.620  & .442 &.596  & .199 &.686  & .567 &.632  & .191 &.642  & .476 &.625  & .191 &.668  & .545 &.628  & .180 &.639  & .460 &.624  & .179 \\
  GS &.619  & .414 &.623  & .184 &.657  & .507 &.622  & .208 &.637  & .437 &.633  & .175 &.690  & .534 &.660  & .169 &.636  & .427 &.634  & .176 \\
 $f_1$*  &.840  & .769 &.868  & .054 &.893  & .883 &.915  & .054 &.783  & .676 &.802  & .073 &.894  & .871 &.926  & .040 &.815  & .720 &.834  & .077 \\
$f$* &.861  & .803 &.887  & .045 &.906  & .899 &.927  & .046 &.815  & .721 &.836  & .060 &.905  & .887 &.933  & .036 &.831  & .743 &.849  & .070 \\
cVAE &.771  & .695 &.842  & .078 &.817  & .812 &.874  & .086 &.747  & .665 &.801  & .085 &.824  & .800 &.895  & .068 &.754  & .659 &.800  & .100 \\ 
Ours &.828  & .747 &.859  & .060 &.860  & .852 &.883  & .071 &.791  & .701 &.816  & .070 &.890  & .878 &.919  & .043 &.810  & .719 &.838  & .070 \\
\bottomrule[1pt]
  \end{tabular}
  \label{tab:model_analysis}
\end{table*}

\textbf{(2) Create one single noisy label for each image:} In previous experiments, our noisy labels are generated by handcrafted feature-based saliency methods in the setting of multiple noisy labels per image. Specifically, we produce three noisy labels for each training image by methods RBD \cite{Background-Detection:CVPR-2014}, MR \cite{Manifold-Ranking:CVPR-2013} and GS \cite{GS_Sal}, respectively. As our method has no constraints on the number of generated noisy labels per image, we conduct experiments to test our models learned in the setting of one noisy label per image. In Table \ref{tab:model_analysis}, we report the performances of the models learned from noisy labels generated by RBD \cite{Background-Detection:CVPR-2014}, MR \cite{Manifold-Ranking:CVPR-2013} and GS \cite{GS_Sal}, respectively. We use $f$-RBD, $f$-MR and $f$-GS to represent their results, respectively. We observe comparable performances with those using the setting of multiple noisy labels per image, which means our method is robust to the number of noisy labels generated from each image and the quality of the generated noisy labels. (RBD ranks the $1^{st}$ among unsupervised saliency detection models in \cite{MSRA10K}. RBD, MR and GS represent different levels of qualities of noisy labels). We also show in Table \ref{tab:model_analysis} the performances of the above handcrafted feature-based methods, which are denoted by RBD, MR and GS, respectively. The big gap between RBD/MR/GS and $f$-RBD/$f$-MR/$f$-GS demonstrates the effectiveness of our model.

\textbf{(3) Train the model from clean labels:} 
The proposed noise-aware encoder-decoder can learn from clean labels, because clean label can be treated as a special case of noisy label, and the noise generator will learn to output zero noise maps in this scenario. We show experiments on training our model from clean labels obtained from the DUTS training dataset. The performances denoted by $f$* are shown in Table \ref{tab:model_analysis}. For comparison purpose, we also train the encoder-decoder component without the noise generator module from clean labels, whose results are displayed in Table \ref{tab:model_analysis} with a name $f_1$*.  
We find that (1) 
our model can still work very well when clean labels are available, and (2) $f$* achieves better performance than $f_1$*, indicating that even though those clean labels are obtained from training dataset, they are still ``noisy'' because of imperfect human annotation. Our noise-handling strategy is still beneficial in this situation.

\textbf{(4) Train the model by variational inference:}
In this paper, we train our model by alternating back-propagation algorithm that maximizes the observed-data log-likelihood, where we adopt Langevin Dynamics to draw samples from the posterior distribution $p_{\theta}(Z|Y,X)$, and use the empirical average to compute the gradient of the log-likelihood in Eq.(\ref{update_phi}). One can also train the model in a conditional variational inference framework \cite{sohn2015learning} as shown in Eq. (\ref{eq:VAE}). Following cVAE \cite{sohn2015learning}, we design an inference network $p_{\phi}(Z|Y,X)$, which consists of four cascade convolutional layers and a fully connected layer at the end, to map the image $X$ and the noisy label $Y$ to the $d=8$ dimensional latent space $Z$. The resulting loss function includes a reconstruction loss $\|Y_i-f(X_i,Z_i,\theta)\|^2$, a KL-divergence loss $\text{KL}( p_{\phi}(Z|Y,X) \Vert p_{\theta}(Z|Y,X) )$ and the edge-aware smoothness loss presented in Eq.(\ref{smoothness_loss}). We present the cVAE results in Table \ref{tab:model_analysis}. Our results learned by ABP outperforms those by cVAE. The main reason lies in the fact that the gap between the approximate inference model and the true inference model, \ie, $\text{KL}( p_{\phi}(Z|Y,X) \Vert p_{\theta}(Z|Y,X))$, is hard to be zero in practise, especially when the capacity of $p_{\phi}(Z|Y,X)$ is less than that of $p_{\theta}(Z|Y,X)$ due to an inappropriate architectural design of $p_{\phi}(Z|Y,X)$. On the contrary, our Langevin Dynamics-based inference step, which is derived from the model, is more natural and accurate.

\section{Conclusion}
Although clean pixel-wise annotations can lead to better performances, the expensive and time-consuming labeling process limits the applications of those fully supervised models.
Inspired by previous work \cite{Zhang_2017_ICCV,Zhang_2018_CVPR,DeepUSPSDR},
we propose a noise-aware encoder-decoder network for disentangled learning of a clean saliency predictor from noisy labels. The model represents each noisy saliency label as an addition of perturbation or noise from an unknown distribution to the clean saliency map predicted from the corresponding image. The clean saliency predictor is an encoder-decoder framework, while the noise generator is a non-linear transformation of a Gaussian noise vector, in which the transformation is parameterized by a neural network. Edge-aware smoothness loss is also utilized to prevent the model from converging to a trivial solution. We propose to train the model by a simple yet efficient alternating back-propagation algorithm \cite{ABP_AAAI,xie2019learning}, 
which is superior to variational inference.
Extensive experiments conducted on different benchmark datasets demonstrate the effectiveness and robustness of our model and learning algorithm.

\small{\vspace{.1in}\noindent\textbf{Acknowledgments.}\quad
This research was supported in part by the Australia Research Council Centre of Excellence for Robotics Vision (CE140100016).}


%
%
\bibliographystyle{splncs04}
\bibliography{saliency_bib}
\end{document}